\documentclass[a4paper, 10 pt, conference]{ieeeconf}

\IEEEoverridecommandlockouts                              % This command is only needed if 
\usepackage{xcolor}

\definecolor{MyDarkBlue}{rgb}{0,0.08,1}
\definecolor{MyDarkGreen}{rgb}{0.02,0.6,0.02}
\definecolor{MyDarkRed}{rgb}{0.8,0.02,0.02}
\definecolor{MyDarkOrange}{rgb}{0.40,0.2,0.02}
\definecolor{MyPurple}{rgb}{111,0,255}
\definecolor{MyRed}{rgb}{1.0,0.0,0.0}
\definecolor{MyGold}{rgb}{0.9,0.8,0.0}
\definecolor{MyDarkgray}{rgb}{0.66, 0.66, 0.66}

\usepackage{graphics} % for pdf, bitmapped graphics files
\usepackage{epsfig} % for postscript graphics files
\usepackage{mathptmx} % assumes new font selection scheme installed
\usepackage{times} % assumes new font selection scheme installed
\usepackage{amsmath} % assumes amsmath package installed
\usepackage{amssymb}  % assumes amsmath package installed
\usepackage{color}
\usepackage{gensymb}
\usepackage{siunitx}

\newcommand{\myparagraph}[1]{\vspace{0.1in}\noindent\textbf{#1}}

\title{\LARGE \bf GelSight EndoFlex: A Soft Endoskeleton Hand with Continuous High-Resolution Tactile Sensing

}

\author{
    \authorblockN{Sandra Q. Liu$^{*}$, Leonardo Zamora Yañez$^{*}$, Edward H. Adelson} % <-this % Leonardo Zamora Yañez
        \authorblockA{Massachusetts Institute of Technology\\
        \authorblockA{$^*$Authors with equal contribution}
    {\tt\small sqliu@mit.edu, lzamora@mit.edu, adelson@csail.mit.edu} % lzamora@mit.edu
    } } 

\usepackage{multirow}
\usepackage{comment}

\begin{document}

\maketitle
\thispagestyle{empty}
\pagestyle{empty}

%%%%%%%%%%%%%%%%%%%%%%%%%%%%%%%%%%%%%%%%%%%%%%%%%%%%%%%%%%%%%%%%%%%%%%%%%%%%%%%%
\begin{abstract}
%Ted's version
We describe a novel three-finger robot hand that has high resolution tactile sensing along the entire length of each finger. The fingers are compliant, constructed with a soft shell supported with a flexible endoskeleton. Each finger contains two cameras, allowing tactile data to be gathered along the front and side surfaces of the fingers. The gripper can perform an enveloping grasp of an object and extract a large amount of rich tactile data in a single grasp. By capturing data from many parts of the grasped object at once, we can do object recognition with a single grasp rather than requiring multiple touches. We describe our novel design and construction techniques which allow us to simultaneously satisfy the requirements of compliance and strength, and high resolution tactile sensing over large areas.

\end{abstract}

%%%%%%%%%%%%%%%%%%%%%%%%%%%%%%%%%%%%%%%%%%%%%%%%%%%%%%%%%%%%%%%%%%%%%%%%%%%%%%%%
\section{INTRODUCTION}

% \tednote{Test comment}
% \slnote{Test comment}
% \lznote{Test comment}

% As the field of soft robotic manipulators continues to grow, we continue to draw inspiration from the seemingly simple complexity of our own human hands. Not only can human hands perform many everyday tasks, they are also able to leverage both the rigidness and softness inherent with bone covered in flesh and muscles. Even though the rigidity of our bones provide hands with both stiffness and strength, the compliance of our flesh allows us to comply with environments around us and interact using this softness. Furthermore, our skin also houses mechanoreceptors, which gives us the ability to perform object recognition with a single grasp \cite{mechanoreceptors}. Being able to perform these tasks allow us, and robots, to solve many useful manipulation problems such as helping the elderly recognize and safely grab a glass of water in a dark room or dig through a bag to help retrieve a book or soft fruit. 

% Ted's version
The human hand has provided inspiration for many robot hands. Human fingers contain an interior articulated skeleton, which is covered with soft skin, providing the fingers with a combination of strength and compliance. The fingers are rounded, with tactile sensing present throughout the skin, and with the best tactile acuity on the front surfaces. When a person holds an object with an enveloping grasp, the object touches the hand at a great many points, allowing the person to recognize the object by its shape, size, and other properties. Our goal is to create a robotic hand that emulates many of these properties.

\begin{figure}[ht!]
	\centering
	\includegraphics[width=1.0 \linewidth]{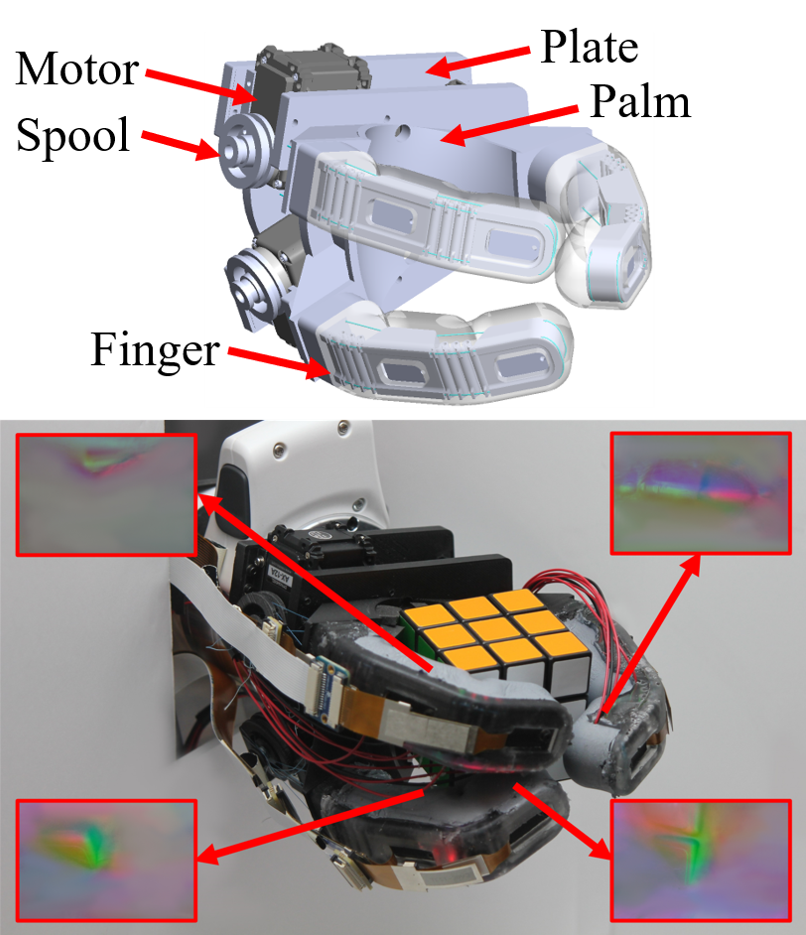}
    \vspace{-15pt}
	\caption{\textbf{Top} A CAD model of our GelSight EndoFlex gripper with some of the parts labeled. \textbf{Bottom} The GelSight EndoFlex is securely grasping a Rubik's cube and the corresponding processed difference images of four of the six sensing regions are displayed. Of note is that the bottom two sensor images are showing continuous sensing along the side and corner of the cube, while the top two sensor images are showing one image each from the other two fingers.}
	\label{fig:bigpic}
	\vspace{-15pt}
\end{figure}

The ability to identify an object using a single grasp is important and requires ``complete" sensing along the grasping surfaces of a finger. Even though many current finger-inspired sensors can perform object recognition well with high-resolution finger tip sensors or with low-resolution larger tactile sensors, they either require that the object be in full contact with the finger tips or multiple regrasps to classify the object within the hand \cite{biotac_rec, palm_sense}. Furthermore, they do not have the compliance afforded by soft robotics, which can greatly improve secure grasping abilities of the gripper or make them safer for interaction with the world around them. 

In other words, soft robotic manipulators could greatly benefit from having structural compliance and rigidity, along with high-resolution sensing of tactile sensors. To this end, we present the following contributions:

\begin{itemize}
    \item A novel design of a continuous high-resolution tactile sensor along a curved surface;
    \item An endoskeleton finger design for a human-inspired gripper that incorporates tactile sensing (Fig. \ref{fig:bigpic});
    \item A neural net that can utilize only the tactile images from a single grasp to classify objects. 
\end{itemize}

\section{RELATED WORK}
\subsection{Hand Grippers}
Human hand inspired grippers have been previously designed with varying degrees of sensing, rigidness and anthropomorphism \cite{tavakoli2014flexture1, xiong2016rigid, choi2006rigidtact}. Although robotic systems were historically composed of rigid materials, interest in soft systems has quickly risen \cite{fumiya2011challenges}. Rigid hands traditionally focused on control systems and force transmission while neglecting contact rich sensing and compliant gripping that more closely characterizes human hands \cite{NovelSoftRobotics}. Soft robotics offers the advantages of compliance, robustness, and can be compatible with high-resolution geometry sensing with camera-based sensors \cite{OGGelSight}.

Rigid robots have often enjoyed well defined kinematic models and high strength, making them ideal manipulators for repeatable and complex motions \cite{CableRigidPowerGrasp}. However, gripping often introduces a degree of uncertainty that may require a softer touch to avoid high energy collisions \cite{rus2015soft}. Soft robotic grippers benefit from their natural robustness and compliance which have proven to be critical when grasping \cite{shintake}. Due to their compliant nature, soft robots are considered to have an infinite degrees of freedom leading to challenges when developing a robust control system. However, recent advances in simulation and robotics have led to the BCL-26, a soft gripper with 26 controllable degrees of freedom that is capable of dexterous motion with a high degree of anthropomorphism. \cite{26DOF}. Other modern designs such as the RBO Hand 3, show great promise with their dexterous manipulation and potential to incorporate sensors due to its larger size \cite{RBOHand3}. 

Many attempts have been made at marrying soft and rigid robotics to achieve flexible yet strong robots \cite{ManufacturingHybridRobotics, uppalapati2020berry, PushPuppet}. One approach to strengthening and increasing precision of soft grippers has been embedding skeletons within their structure \cite{BernTendonHand}. Although the addition of an endoskeleton brings various benefits, it also comes with some drawbacks including increased manufacturing and modeling complexity. To combat the increase in complexity, simulation has become a popular tool to supplement control design \cite{HybridRoboticsFEA}. The properties of soft-rigid robotics appear to be a significant step towards high fidelity biomimetic hand grippers. 

Despite the various advances in robotics to achieve a soft human-like hand, there are still critical elements missing from current designs. Most notably, there is an absence of rich geometric-based sensing in rigid and soft hands alike \cite{CableRigidPowerGrasp, 26DOF}. Therefore, there is still progress to be made in developing a soft anthropomorphic hand with geometry sensing capabilities.

\subsection{Sensing and Soft Grippers}

Most previous tactile sensing work in robotic grippers has been force-based using capacitive or strain sensors \cite{CapacitiveSkinFingers,UnderwaterRobotStrainGauge}. These sensors provide a low cost option with high response time, but these types of sensors are better for sensing stiff and flat surfaces \cite{ForseSensorFlatSurafces}. Vision-based sensors can provide additional sensing data and be highly compatible with soft robots.

Existing vision-based systems rely on cameras to capture the deformation of some elastomer and process the footage to obtain tactile data \cite{OGGelSight}. One such sensor is the TacTip which uses a camera to measure the deformation of a silicone membrane and superresolution to achieve precise force localization \cite{OGTacTip}. The soft nature and highly accurate sensing of TacTip has great potential, but the sensor size and lack of geometry sensing limit its application to anthropomorphic hands. The GelSight sensor family offers an alternative with its high resolution tactile sensing and application to curved surfaces \cite{RomeroCurvedTip}. GelSight sensors operate with a camera that views a painted aluminum-silicone skin that can capture finely detailed tactile imprints on its surface. This surface is then illuminated by different LEDs.

Previous GelSight sensing area has been limited by uni-camera sensing, wide angle lenses with some distortion, and their large size \cite{wedge,ExoGelSight,GelSightHand,LiuFinRay}. GelSight applications have seen limited integration of the sensing surface with the gripper body \cite{CableManipulationGS}. Therefore, there is still space to explore soft human-like grippers with structural integration of tactile sensors. One potential design for extending sensing surface area is to expand on the work of She \emph{et al.} \cite{ExoGelSight} by using two or multiple cameras to create a continuous sensing surface. To our knowledge, no other GelSight sensor has used multiple cameras to create one continuous and compact sensing surface. Our novel design provides wide range GelSight sensing in a compact and soft anthropomorphic package.

% -Conclude with how we incorporate human hands an sensing into one gripper 

% - maybe also talk about multiple touches to recognize an object/do classification, whereas continuous sensing will allow us to do a single touch (intro OR related works)
 
 %applications of ML??? 
 
 %\cite{RigidHandForceSensing}. Other grippers have well defined kinematic models but contain a rigid exoskeleton, sacrificing anthropomorphism \cite{CableRigidPowerGrasp}.
 
%  Most soft gripper with complex motion often omit accurate pose estimation This wave of soft grippers also come with the challenge of accurate position estimation [ST]. 

%-To the knowledge of the author, this is the first finger to combine high resolution tactile sensing in an anthropomorphic package 

% The dexterity of a human hand with the rich sensing of GelSight!

% Sensing Grippers
% - Methods of Sensing
% - Force Sensors
% - GelSight
% - Optical cables
\section{METHODS}

\subsection{Hardware}
\begin{figure}[ht]
	\centering
	\includegraphics[width=1.0 \linewidth]{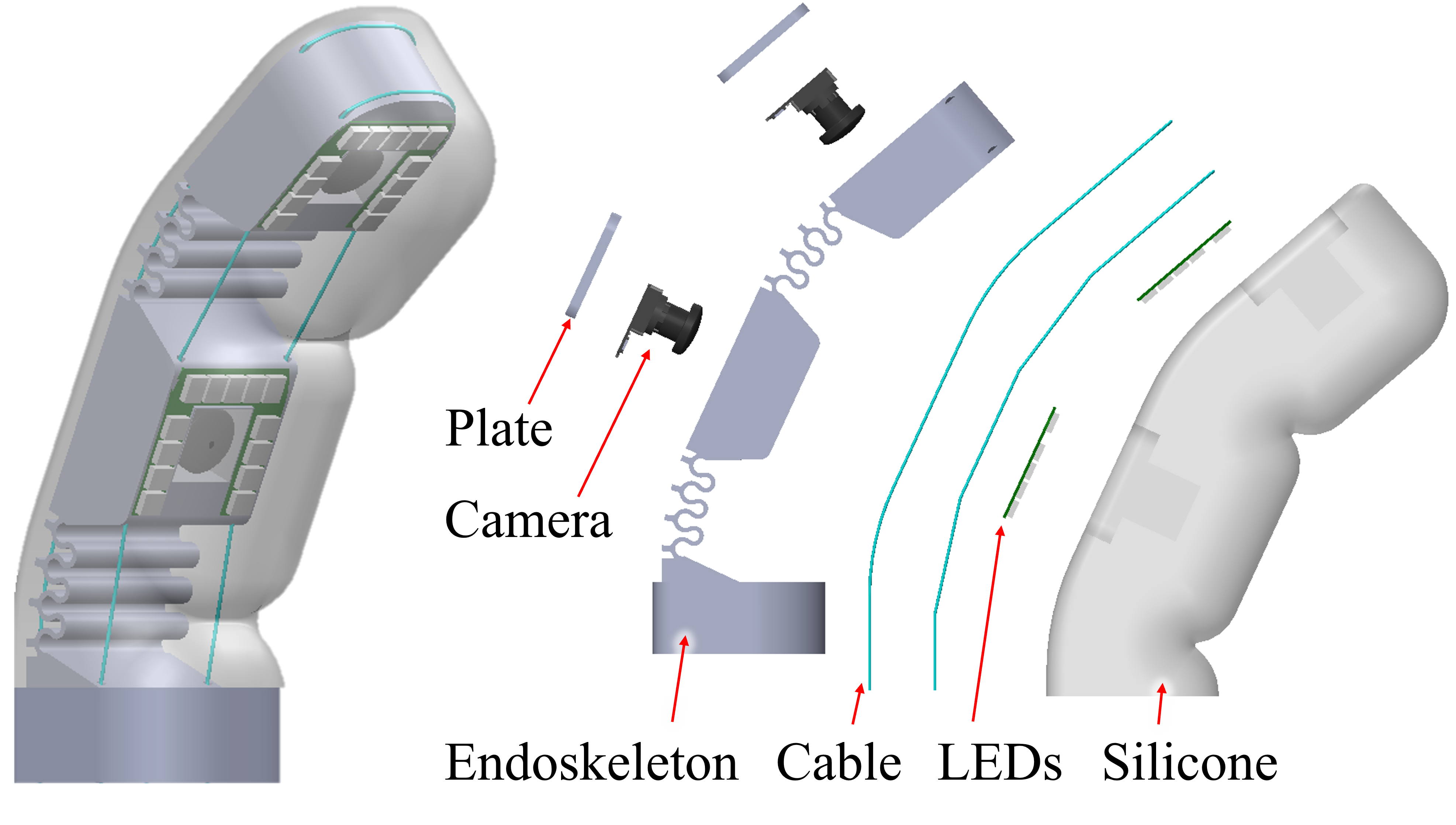}
    \vspace{-20pt}
	\caption{A close-up view of an EndoFlex finger with an exploded view. Each finger operates independently with one degree of freedom and can be quickly replaced if damaged.}
	\label{fig:explode}
\end{figure}

The EndoFlex sensor is composed of an endoskeleton encased in silicone with two embedded cameras for continuous sensing (Fig. \ref{fig:explode}).  Each endoskeleton was designed to be one continuous piece with a pair of rigid segments and flexures to form joints. This design minimizes the number of parts required to fabricate one finger when compared to traditional rigid fingers. The flexure design was chosen for its high compliance and low deformation of individual elements to reduce silicone delamination. We 3D printed the endoskeleton using an Onyx One printer with Markforged Onyx plastic for its combination of high strength and relatively low tensile modulus when compared to other extruded plastics. This combination of properties allowed minimal force loss during actuation. 

A camera was mounted into each endoskeleton segment to prevent any shifting during actuation. Three sets of red, green, and blue LEDs were mounted with cyanoacrylate adhesive onto the rigid segment of the endoskeleton. They were spaced 90 degrees apart to create a colored light gradient for the GelSight algorithm. Finally, the endoskeleton was threaded with Piscifun Onyx Braided Fishing Line soaked in Smooth-On Universal Mold Release to reduce friction when cast in silicone. We chose to use cable-driven actuation to reduce potential camera-view obstructions and also so that we could more easily integrate the camera into the finger skeleton.

A rigid three finger palm was designed with temporary fasteners to allow for fast replacement of damaged fingers or for future iterations. Fingers were positioned in a `Y' pattern with two fingers and an opposing thumb. The pair of fingers was spaced thirty degrees apart to distribute grasping force without creating collisions. The palm was designed to have a rounded feature with a polyurethane foam layer to add grasping ability. A separate rigid plate was designed to be fastened onto the Panda robotic arm. Three Dynamixel AX-12A servos were mounted between the plate and the palm and served as the actuation method for the fingers through double axle spools. The double axle design allowed for actuated contraction and extension of each finger. The palm, plate and spools were all printed with Markforged Onyx plastic using a Markforged Onyx printer. 

\begin{figure}[ht]
	\centering
	\includegraphics[width=1.0 \linewidth]{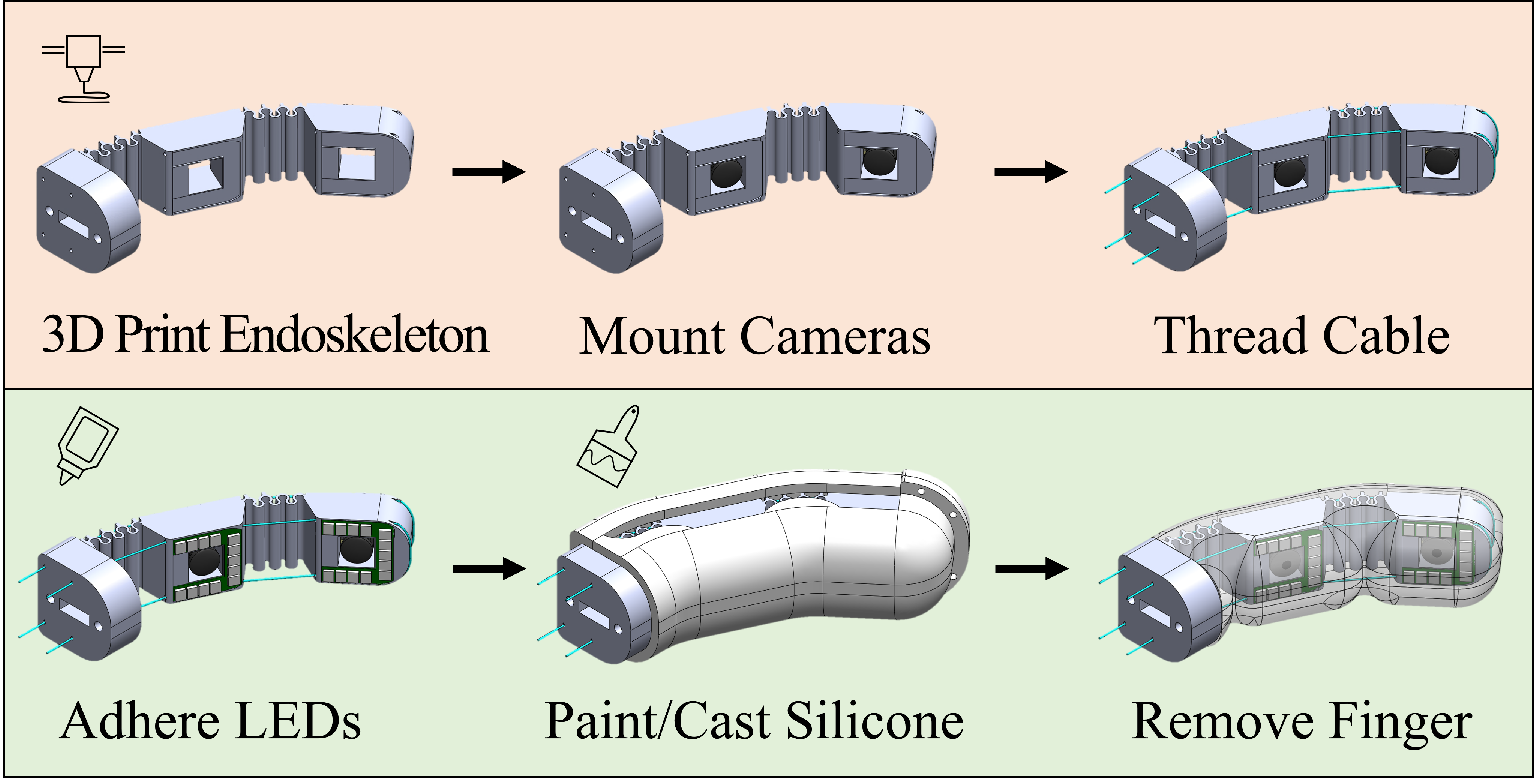}
    \vspace{-20pt}
	\caption{The manufacturing process for the EndoFlex sensor including assembly of electronics and casting of silicone.}
	\label{fig:manu}
\end{figure}

As part of our finger manufacturing process, which is fully shown in Fig. \ref{fig:manu}, a two part mold is designed for casting silicone to create the optically clear medium for the GelSight sensor. The mold was designed to hold the endoskeleton during the casting process which removed the need for fasteners or adhesives to hold the silicone layer. The mold had high curvature to create a rounded finger much like a human finger. One major benefit of the curved surface was the high reflection of lights within the silicone which aided in sensing by removing shadows of pressed objects. The mold design removed any air gap between the camera lens and cast silicone to minimize the refraction of light. The mold was produced using a Formlabs 2 SLA printer for its high resolution. To achieve the optical clarity required to use GelSight, the mold was incrementally sanded with sandpaper reaching 2000 grit.  

To allow the silicone to compress when the tendons pulled the endoskeleton finger to a closed grasp position, we chose to synthesize a softer silicone for the finger. As a result, we used a ratio of 1 to 15 to 5 parts of XP-565 parts A and B, and plasticizer (Phenyl Trimethicone, Lotioncrafter). Decreasing the ratio of part A to B for the XP-565 is equivalent to adding less catalyst, which increases the softness of the silicone, while the addition of the plasticizer also causes the resulting cured silicone to have a softer texture. 

Before pouring the silicone mixture into the mold, we used a paint brush to paint a thin layer of Inhibit-X (Smooth-On Inc). After waiting a few minutes for it to dry, we sprayed a layer of Ease Release 200 (Smooth-On Inc) on the mold. To create the sensing surface, we combined 2.5 parts 4 $\mu$m Aluminum cornflakes (Schlenck) with a mixture of 11 parts silicone ink catalyst and gray silicone ink base (Raw Materials Inc.) and 30 parts NOVOCS Gloss (Smooth-On Inc), and mixed it for a minute using an ultrasonicator. This mixture was then sprayed into the inside of the top mold with an airbrush and left to dry for at least 10 minutes before we fit the threaded endoskeleton inside of the mold and screwed the mold halves together. Remaining holes and the lips of the mold were covered in a thin layer of brushed-on silicone adhesive (Devcon), which created a seal for the mold and prevented any silicone leakage outside of the mold that could be caused by mold warping or other printing imperfections. 

Once the main body silicone mixture had been degassed, we slowly poured the mixture into the prepared mold. The entire mold assembly was placed on top of a vibrating plate for 10 minutes to get rid of any bubbles in the camera-viewable areas. These bubbles may have been induced by the silicone pouring over the flexures, electronics, and other 3D printed parts inside of the mold. Some of the bubbles were retained along the side of the sensor surface, which is not viewable by the camera and did not negatively affect the sensor integrity. 

\begin{figure}[ht]
	\centering
	\includegraphics[width=0.8 \linewidth]{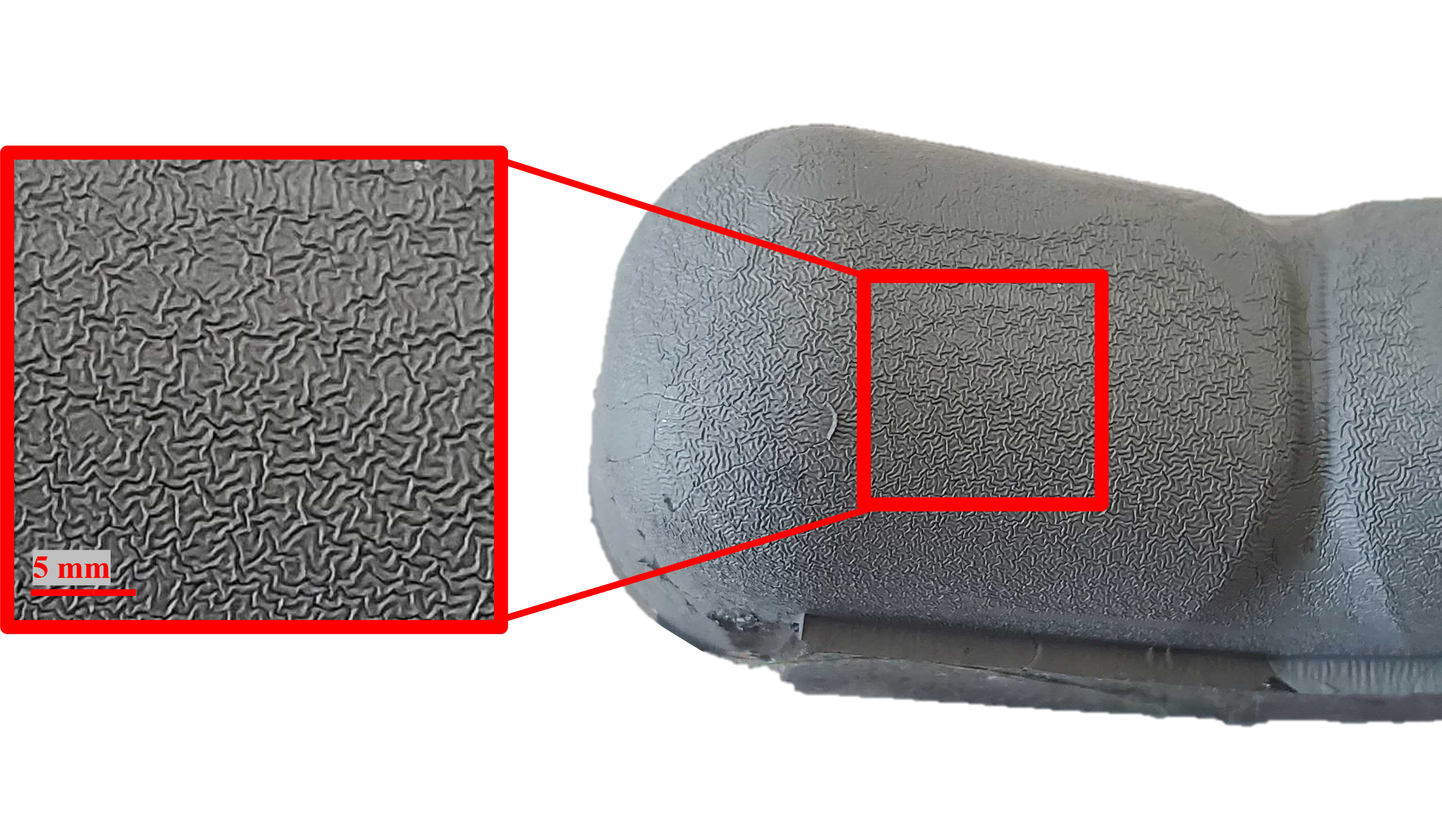}
    \vspace{-20pt}
	\caption{A close up image of the reticulated wrinkle surface of the GelSight EndoFlex sensor. The width of one of the wrinkles is approximately 0.4 mm wide and was only created when we first sprayed the paint on the mold surface before casting silicone inside.}
	\label{fig:wrinkle}
\end{figure}

Finally, the mold was placed inside of a oven at 125$^{\circ}$F (52$^{\circ}$C) for 12 to 15 hours. This temperature was chosen to prevent any of the electronics or inner structures from reaching their glass temperatures and causing delamination of the parts from the silicone. Once the finger was removed from the mold, the gray sensing membrane surface was no longer smooth and instead had a reticulated wrinkled texture (Fig. \ref{fig:wrinkle}). This phenomena only occurred when we sprayed the paint on the mold first and did not occur if we chose to cure the finger first without the paint in the mold and spray the paint on the finger surface afterwards. 

The modular fingers were then placed on our palm plate to create our completed gripper. We also note that this configuration can be changed to enable different types of grasps, although we chose an enveloping grasp to maximize the amount of sensing the gripper could obtain from grasping an object in its palm. 

\subsection{Software}
\begin{figure}[ht]
	\centering
	\includegraphics[width=1.0 \linewidth]{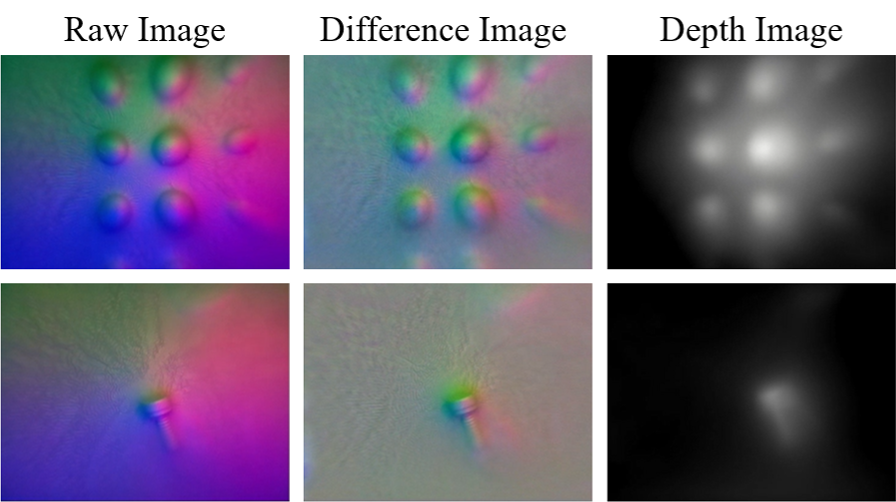}
    \vspace{-20pt}
	\caption{From left to right, we have raw sensor images of a 3.75 mm ball bearing array and a M2 screw, followed by their difference images from a reference image (no tactile contact), and the corresponding uncalibrated depth image.}
	\vspace{-10pt}
	\label{fig:diff}
\end{figure}

Each finger was equipped with two Raspberry Pi Zero spy cameras with a 160$^{\circ}$ field of view, for a total of six cameras. All of the cameras were able to view a curved segment of the finger, which was illuminated by tri-directional LEDs. The finger segment images were individually streamed using the mjpg-streamer package and can be processed using OpenCV and fast poisson solver \cite{opencv, poisson} to get difference images and uncalibrated reconstruction images, as shown in Fig. \ref{fig:diff}.

\section{EXPERIMENT}
To show the usefulness of having continuous sensing, we collected single grasps of various objects and performed a classification task based on the entire finger sensing region. Previous works show that object classification using finger tip sensing or low-resolution palm sensing is accurate, but only when the objects were in contact with the fingertips or multiple touches have been performed \cite{biotac_rec, palm_sense}. 

\begin{figure}[ht]
	\centering
	\includegraphics[width=1.0 \linewidth]{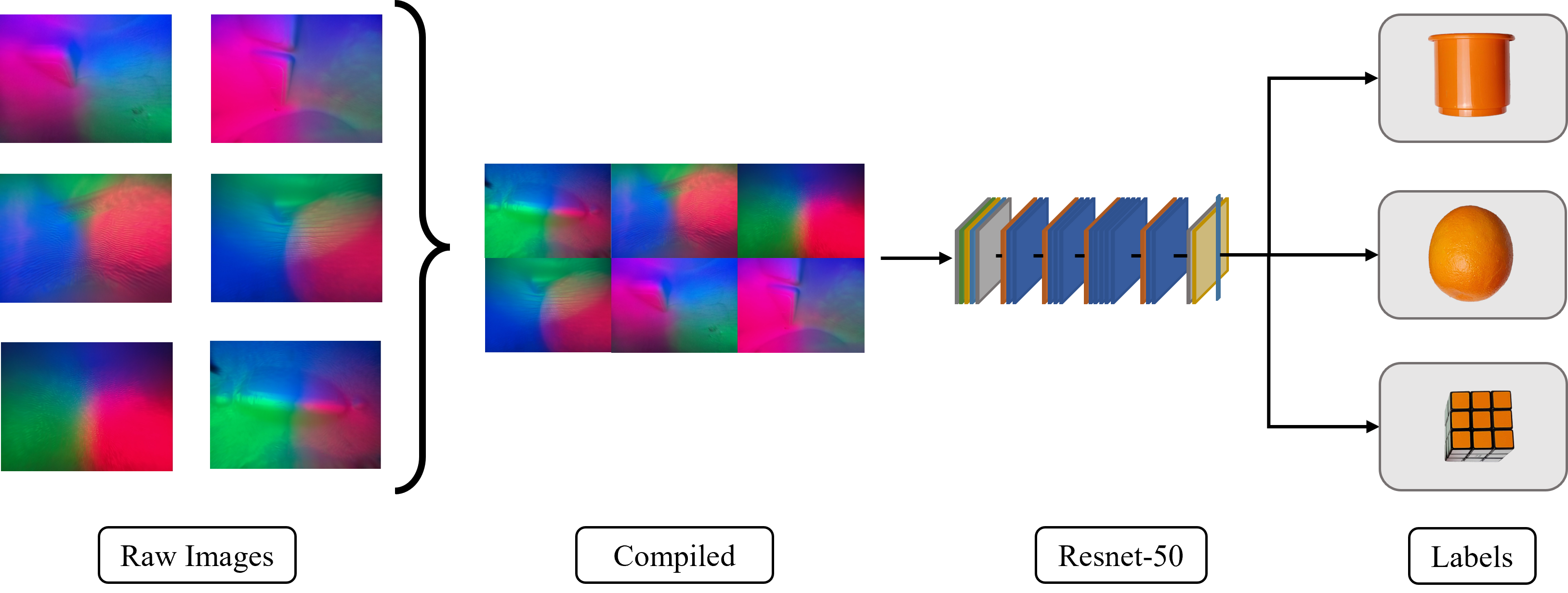}
    \vspace{-20pt}
	\caption{Neural net architecture for our single grasp classification. Once the object has been grasped, the six images are stitched together in a 2x3 array, thrown into our Resnet architecture and classified into a toy cup, an orange, or a Rubik's cube.}
	\label{fig:resnet}
\end{figure}

Our grasping object set included three distinct objects from the YCB dataset: the Rubik's cube, one of the toy stacking cups, and a plastic orange \cite{YCB}. These three objects are shown in Fig.  \ref{fig:resnet}. For each object, we collected approximately 500 different grasps using all six of the cameras inside the fingers to obtain a holistic, ``full-hand" tactile view of the entire object. To capture many different grasps, we had assistants manually reorient each object randomly such that it could still be feasibly grasped with the gripper, which allowed different parts of the sensor images to capture different features of the object that was being grabbed. We also attempted grasps utilizing a couple of the fingers instead of all of the fingers in the cases that the third finger did not have a solid contact with the object in its hand.

For each set of six images we captured, we stitched them together into a 2 by 3 array and used them as inputs for a Resnet-50 neural net architecture with the three outputs as the objects we used for our grasping data set \cite{resnet}. We chose to use stochastic gradient descent as our optimizer, with a learning rate of 1e-3 and a learning rate scheduler with a step size of 7 and a gamma set to 0.1. We also implemented data augmentation on the entire set of images to deal with potential inconsistent lighting or random noise output of the images, and to account for eventual wear and tear in the silicone over time. We split our data into training and validation sets in a 80\% to 20\% ratio. The complete neural net architecture is shown in Fig. \ref{fig:resnet}.

\subsection{Results}

\myparagraph{Grasping} 
The GelSight EndoFlex was able to easily and very securely grasp all of the objects in our object set. In particular, the polyethylene foam layer on the palm provided a compliant, deformable surface that the grasped objects could be pressed against. The hand was also able to grasp empty water bottles without crushing them, as well as heavier objects, like a drill with a battery, without dropping them. As expected, the compliance of the soft gel allowed us to grasp more fragile objects, while the rigid endoskeleton allowed the fingers to withstand the force and weight of a heavier object.

Each finger was also able to bend to around 60$^{\circ}$ at each flexure point using the Dynamixel motors. Because the silicone was quite soft and because we added human finger-inspired grooves along the flexures, when the fingers bent, the silicone was able to more easily compress around the sides. However, the silicone still obstructed some of the bending angle, and as a result, the endoskeleton finger was unable to bend to its full 90$^{\circ}$ range that it would have been able to otherwise. Furthermore, deepening the grooves to facilitate bending would have limited the sensing area and ultimately interfered with the continuous sensing. Nonetheless, this limitation in motion did not severely limit the hand's ability to grasp objects because the deformable silicone surface over the endoskeleton finger helped to accommodate any loss of motion with its compliance and softness. 

Casting the mold while the finger was in a slightly bent position helped to prevent creases in the surface of the silicone when the finger bent. Doing so also prevented silicone creasing when the finger was straightened out since the sensing surface was pulled in tension. Unfortunately, over time, pulling the silicone finger in tension caused parts of the silicone in the base of the finger to slightly tear. We believe that this problem could potentially be mitigated by using a softer silicone with higher elongation. 

Finally, given a different arrangement of fingers or with a finger that could behave more thumb-like with an added degree of freedom, we believe that these fingers have the potential to grasp an even larger variety of objects.

\myparagraph{Tactile Sensing}
As designed, the finger was able to continuously sense along the entire length of the finger when it was in a ``closed" position. The fingers were also able to sense along the sides as well, although some sensing was slightly lost at the very tips of the fingers. 

Overall, the finger was able to provide extremely high resolution sensing and the raw sensor images were able to capture details that previous GelSight sensors could sense, but with additional sensing coverage due to its rounded shape and the wider camera field of view. However, the wider camera field of view and the curved shape caused some distortion in the sensing image, which is most apparent on the sides of the image frame. 

Additionally, some of the sensing surfaces appeared to have distinct rings of lights around the different color channels instead of the blending we would have expected from using a Lambertian paint on the surface of the silicone gel. We believe that this phenomena could have been caused by slight delamination of the silicone from the LEDs. The addition of the air interface will cause the light to refract from the air to the silicone face and potentially cause these rings of light to form and prevent even blending of the light within the silicone. In particular, we noticed that when objects were pressed against these sensing surfaces, the light circles began to dissipate. Nonetheless, this did not affect the sensor resolution and the distinct features of the objects were still distinguishable as the tactile sensor had extremely high-resolution.

Finally, we noticed that the wrinkles, which were manufactured on some of the finger sensing surfaces, were helpful in preventing tears in the silicone membrane. Unlike the smoother sensing surfaces, it seems like the wrinkles helped to mitigate the high stress points caused by sharp corners poking into the sensor surface. The surfaces with wrinkles also felt like they had less friction than the smoother surfaces. Although the wrinkled surface made surface reconstruction difficult because the wrinkled texture appeared in difference images, they did not seem to negatively affect our object classification. The effect could also have been mitigated since we noticed that if enough pressure was put on the sensor surface, the wrinkles would smooth out slightly, which would not affect object classification results. 

\myparagraph{Object Classification}
Our object classification model was able to obtain 94.1\% accuracy on our validation set. In live testing, which consisted of our robotic hand grasping the 3 objects ten times each, we were able to correctly classify 80\% of the objects. The orange was able to be recognized 9 out of 10 times, while the classifier slightly struggled with distinguishing between the Rubik's cube and the toy cup (80\% and 70\% accuracy, respectively). We believe that the discrepancy in the validation set results and the live testing results could be due to slight tears that developed over the course of the data collection and testing. Regardless, the hand was able to only use a single grasp to recognize the identity of an object. 

As we expected, the orange, which had the most distinguishable tactile features, was the easiest for our model to recognize. Not only was the orange covered in an unique bumpy skin texture, it also had a distinctive stem portion. On the other hand, unless the fingers directly pressed against a corner of the Rubik's cube or along multiple smaller cubes, it was hard to visually distinguish some of its edges from the edges at the bottom and top of the toy cup.

We believe that this confusion between the Rubik's cube and the toy cup could be mitigated by adding a palm, which could also provide additional sensing. The added sensing from a larger area on the palm could have helped capture more tactile details that may have been missed by the fingers. Regardless, the object classification using continuous sensing along the multi-fingered hand was fairly robust and able to perform well on our object set. Specifically, it could be useful for grabbing objects in the dark or in an occluded environment where external vision would not be useful or could not be used.

\section{CONCLUSION AND DISCUSSION}

In this paper, we present the novel design of a continuous high-resolution tactile sensor incorporated into a finger, which was then integrated into a human-like hand. The hand was then able to use these large sensing ranges to be able to somewhat accurately classify objects using a single grasp, which, to the authors' knowledge, has not been done before. The ability to identify an object with a single grasp is akin to the way we as humans are able to grab an object with some priors and without external vision and determine almost immediately what we are holding. 

Although recent research has focused a lot on large range low-resolution tactile sensors or high-resolution fingertip sensors for dexterous manipulation, not much research has been done on high-resolution sensing across the majority of a finger's surface. Having this added sensing allows us to perform many useful classification tasks, and doing so in a soft, compliant gripper allows us to also safely and securely interact with objects and the surrounding environment. Sensors similar to the GelSight EndoFlex have the ability to be used for home-care robots or for human-robot interaction, where compliance and sensing are key to success. 

Future work on this gripper involves adding a thumb-like joint, as well as full fingertip sensing, which can greatly improve the usability of the gripper for sensing and dexterous manipulation tasks. We can also continue to draw inspiration from GelSight sensors and add markers which could help track slip and shear or torsional forces along the surfaces of the finger. Overall, our novel endoskeleton finger design begins to solve the problem of designing human-inspired soft-rigid robotic hands with high-resolution sensing that are capable of performing more and more complicated tasks.

\section{ACKNOWLEDGEMENTS}

This work was supported by funds from the Toyota Research Institute, the Office of Naval Research, and the SINTEF BIFROST (RCN313870) project. The authors would also like to thank James M. Bern and Megha H. Tippur for their helpful advice and design tips. 

\addtolength{\textheight}{-0cm}   % This command serves to balance the column lengths



\bibliographystyle{IEEEtran}
\bibliography{Ref}

\end{document}